\documentclass[acmsmall]{acmart}

\usepackage{times}
\usepackage{epsfig}
\usepackage{graphicx}
\usepackage{amsmath}
\usepackage{bm}
\usepackage{booktabs}
\usepackage{siunitx}

\usepackage{pifont}
\usepackage{enumitem}

\AtBeginDocument{%
  \providecommand\BibTeX{{%
    \normalfont B\kern-0.5em{\scshape i\kern-0.25em b}\kern-0.8em\TeX}}}

\setcopyright{acmcopyright}
\copyrightyear{2018}
\acmYear{2018}
\acmDOI{10.1145/1122445.1122456}

\acmJournal{JACM}
\acmVolume{37}
\acmNumber{4}
\acmArticle{111}
\acmMonth{8}

\begin{document}

\title{Robust Cross-view Gait Recognition with Evidence: A Discriminant Gait GAN (DiGGAN) Approach}

\author{Bingzhang Hu}
\email{bingzhang.hu@ncl.ac.uk}
\orcid{0000-0002-3592-9341}
\affiliation{%
	\institution{School of Computing, Newcastle University}
	\city{Newcastle upon Tyne}
	\country{UK}
	\postcode{NE4 5TG}
}
\author{Yu Guan}
\email{yu.guan@ncl.ac.uk}
\affiliation{%
	\institution{School of Computing, Newcastle University}
	\city{Newcastle upon Tyne}
	\country{UK}
	\postcode{NE4 5TG}
}

\author{Yan Gao}
\affiliation{%
  \institution{Department of Computer Science and Technology, University of Cambridge}
  \city{Cambridge}
  \country{UK}}

\author{Yang Long}
\email{yang.long@ieee.org}
\affiliation{%
	\institution{Department of Computer Science,Durham University}
	\city{Durham}
	\country{UK}}

\author{Nicholas Lane}
\email{ndl32@cam.ac.uk}
\affiliation{%
	\institution{Department of Computer Science and Technology, University of Cambridge}
	\city{Cambridge}
	\country{UK}}
\author{Thomas Ploetz}
\email{thomas.ploetz@gatech.edu}
\affiliation{%
	\institution{School of Interactive Computing, Georgia Institute of Technology}  
	\country{USA}}

\renewcommand{\shortauthors}{Bingzhang Hu, et al.}

\begin{abstract}
 Gait as a biometric trait has attracted much attention in many security and privacy applications such as identity recognition and authentication, during the last few decades. Because of its nature as a long-distance biometric trait, gait can be easily collected and used to identify individuals non-intrusively through CCTV cameras. However, it is very difficult to develop robust automated gait recognition systems, since gait may be affected by many covariate factors such as clothing, walking speed, camera view angle \emph{etc}. Out of them, large view angle changes has been deemed as the most challenging factor as it can alter the overall gait appearance substantially. 
 Existing works on gait recognition are far from enough to provide satisfying performances because of such view changes. Furthermore, very few works have considered evidences --- the demonstrable information revealing the reliabilities of decisions, which are regarded as important demands in machine learning based recognition/authentication applications. To address these issues, in this paper we propose a Discriminant Gait Generative Adversarial Network, namely DiGGAN, which can effectively extract view-invariant features for cross-view gait recognition; and more importantly, to transfer gait images to different views -- serving as evidences and showing how the decisions have been made. Quantitative experiments have been conducted on the two most popular cross-view gait datasets, the OU-MVLP and CASIA-B, where the proposed DiGGAN has outperformed state-of-the-art methods. Qualitative analysis has also been provided and demonstrates the proposed DiGGAN's capability in providing evidences.
\end{abstract}



\maketitle
\section{Introduction}
In the past few decades, many kinds of biometric traits such as voice \cite{graves2013speech}, face \cite{schroff2015facenet}, fingerprint \cite{maltoni2009handbook}, iris \cite{wildes1997iris} and gait \cite{guan2014reducing}, have been utilised for recognition and authentication. Although the effectivity (accuracy) has always been a major concern during selecting a proper biometric trait, in practise the collectability is also a considerable factor. For example, the iris recognition can usually obtain high accuracies, but the iris recognition systems typically require the subject to cooperate during scanning, which is apparently not always feasible in some scenarios. Apart from the effectivity and collectability, the reliability (readability) also plays an important role in many scenarios, \emph{e.g.} in the courtroom. Such reliability is usually reflected by the evidences, which are typically provided and judged by domain experts. For example, in a 2005 case \cite{larsen2008gait} in Denmark concerning a bank robbery, the gait experts found the suspect shares the same `unique gait', which includes `restless stance, anterior positioning of the head showing a neck lordosis and inversion of the ankle joint' with the robber, based on which the suspect was found guilty. For intelligent systems, the demonstrable information that reveals the decision making process is also referred to as `evidence' in this paper.


Gait, considering the collectability, its advantages can be concluded in many folds. First, the collection can be conducted in distance, and there are many cameras installed in public places such as airports, government buildings, streets and shopping malls to facilitate this. Secondly, collecting gait information does not require the subjects' cooperations, which is helpful in many scenarios such as public security maintaining, identity recognition and authentication. Lastly, gait is a behavioural biometric trait and does not require high-resolution data for remote individual identification.

\begin{figure}[t]
\begin{center}
\includegraphics[width=.8\linewidth]{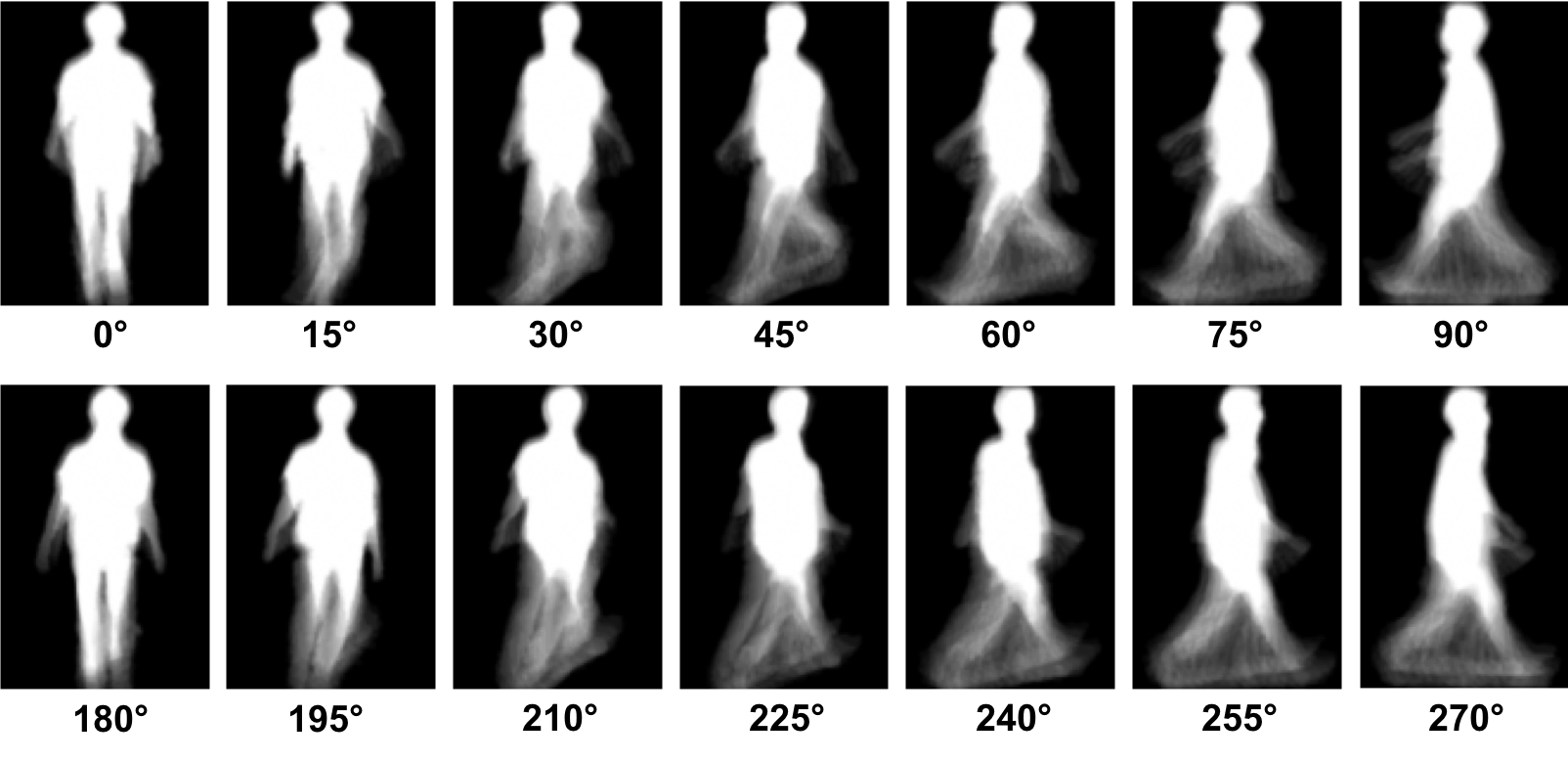}
\end{center}
   \caption{Gait Energy Images (GEIs) in OU-MVLP dataset.}
\label{fig:dataset}
\end{figure}

Considering above superiorities, gait recognition has received much attention from the community during the last few years. However, the covariate factors in gait images such as shoe type, carrying condition, clothing, speed, and camera viewpoint bring many challenges to gait recognition and authentication. Out of them, the large camera viewpoint has been considered as the most challenging factor that may affect the gait features in a global manner. For instance, Fig. \ref{fig:dataset} demonstrates several GEIs\cite{Han_GEI}\footnote{GEI: gait energy image, also know as the average silhouette over one gait cycle, is a popular gait representation proposed in \cite{Han_GEI}. The averagely operation encodes the information of binary frames into a single grey scale image, which makes GEI computationally efficient and less sensitive to segmentation errors.}  in the OU-ISIR Gait Database, Multi-View Large Population Dataset (OU-MVLP) \cite{OU-MVLP2018} from different view angles, and we can see that different views substantially alter the visual features of gaits, causing recognition difficulties. Nevertheless, existing works \cite{wu2017comprehensive, takemura2017input, CCA, TILT, PAMI17_CNN, VTM_TSVD, hu2013view, OU-MVLP2018} on cross-view gait recognition are biased to designing powerful gait representations thus ignore to provide the `evidences' to support their decision making process. Lacking such demonstrable information makes existing cross-view gait recognition techniques far from being used as a modern intelligent system. 


To this end, we propose a Discriminant Gait Generative Adversarial Network (DiGGAN) in this paper. As shown in Fig.~\ref{fig:framework}, the proposed DiGGAN first extracts view-invariant features for cross-view gait recognition tasks, then the view-invariant features will be combined with a user-specified angle condition to generate GEIs in the target views, which may be used as important evidences. Moreover, to enhance the discriminability of the view-invariant features, a local enhancer is introduced to further refine them. Our contributions are summarised as follows:
\begin{itemize}
  \item Framework: A generative adversarial network based model (named DiGGAN) is proposed in this paper. With the mechanisms of two independent discriminators, the proposed network can generate gait energy images (GEIs) at different views while preserving the identity information. 
  \item Performance: On the world's largest OU-MVLP gait dataset \cite{OU-MVLP2018}(with more than 10000 subjects), our method outperforms other algorithms significantly on many real-world cross-view gait identification scenarios (\emph{e.g.}, cooperative/uncooperative mode). It also has the best results on the popular CASIA-B dataset and shows reasonable generalisation ability across datasets. 
   \item Evidence: The proposed DiGGAN is able to generate different view gait evidences, which are important for a wide range of security and privacy applications.
 \end{itemize}
\section{Related Work}
\subsection{Conventional Cross-view Gait Recognition and Identification}
Cross-view gait recognition methods can be roughly divided into three categories.
Methods belonging to the first category (\emph{e.g.} \cite{3d_gait1}) are based on 3D reconstruction through images from multiple calibrated cameras.
However, these methods require a fully controlled and cooperative multiple camera environments, which limits its application in real-world surveillance scenarios.
Methods in the second category perform view normalization on gait features before matching.
In \cite{Goffredo}, after estimating the poses of lower limbs, Goffredo et al. extracted the rectified angular measurements and trunk spatial displacements as gait features.
Kusakunniran et al. \cite{TILT} proposed a view normalization framework based on domain Transformation obtained through Invariant Low-rank Textures (TILT), and gaits from different views are normalised to the side view for matching.
These methods yielded reasonable recognition accuracies in some cross-view recognition tasks, yet they are sensitive for views when gait features are hard to estimate (\emph{e.g.}, frontal/back views).

The third category is to learn the mapping or projection relationships of gaits across views.
The learning process relies on the training data that covers the views appearing in the gallery and probe.
Through the learned metric(s), gaits from two different views can be projected into the common subspace for matching.
In \cite{VTM06}, Makihara et al. introduced the SVD-based View Transformation Model (VTM) to project gait features from one view into another.
After pointing out the limitations of SVD-based VTM, a method in \cite{VTM_SVR} reformulated VTM construction as a regression problem.
Instead of using the global features in \cite{VTM06}, local Region of Interest (ROI) was selected based on local motion relationship to build VTMs through Support Vector Regression (SVR).
In \cite{VTM_SR_CSVT12}, the performance was further improved by replacing SVR to Sparse Regression (SR).

Instead of projecting gait features into a common space, Bashir et al. \cite {CCA} used Canonical Correlation Analysis (CCA) to project gaits from two different views into two subspaces with maximal correlation, and the correlation strength was employed as the similarity measure for identification.
In \cite{motion-co-clustering}, after claiming there may exist some weakly or non-correlated information on the global gaits across views \cite{CCA}, motion co-clustering was carried out to partition the global gaits into multiple groups of gait segments. 
For feature extraction, they performed CCA on these multiple groups, instead of the global gait features as in \cite{CCA}.
Different from most works (\emph{e.g.} \cite{VTM06, motion-co-clustering, CCA}) with multiple trained projection matrices for different view pairs, Hu et al. proposed a novel unitary linear projection named View-invariant Discriminative Projection (ViDP) \cite{ViDP}.
The unitary nature of ViDP makes cross-view gait recognition feasible to be performed without knowing the query gait views.
\subsection{Deep Learning Based Cross-view Gait Recognition and Identification} 
Recently, deep learning approaches \cite{GEINet16, PAMI17_CNN, GaitGAN17, he2019multi} were applied for gait recognition, which can model the non-linear relationship between different views.
In \cite{GEINet16}, the basic CNN framework, namely GEINet was applied on a large gait dataset, and the experimental results suggested its effectiveness when the view angle changes between probe and gallery are small. 
To combat large view changes, a number of CNN structures were studied in \cite{PAMI17_CNN} on the CASIA-B dataset (with 11 views from \ang{0} to \ang{180}), and Siamese-like structures were found to yield the highest accuracies.
However, this dataset only includes $124$ subjects, and the most recent work \cite{OU-MVLP2018} found these CNN structures do not generalise well to a large number of subjects.
It is worth noting that there are some recent methods \cite{AAAI_gaitset, wangliang_TIP, wu2015learning} utilising the silhouette images for gait recognition and identification. For example, Chao et al. \cite{AAAI_gaitset} proposed a framework that employs several weights-shared CNNs for each frame in a silhouette set, and then applies the horizontal pyramid mapping on the feature map to obtain the pooled features for recognition. Zhang et al. \cite{wangliang_TIP} proposed a CNN based framework to extract both spatial and temporal information existing in the gait silhouette, where each frame in a silhouette is first partitioned into several parts and for each part, an independent CNN is applied to extract spatial specific information. Additionally, an LSTM attention model is employed in their framework to obtain attention scores thus yield better temporal information compared to existing methods. However, compared to GEIs, silhouette images suffer from the segmentation noises and generally require more computational resources. Due to this fact, in this work, we will not consider the silhouette-based methods.

As another branch of deep learning technique, Generative Adversarial Networks (GANs) \cite{34_GAN} introduce a novel self-upgrading system. By keeping a balanced competition between a generator and a discriminator, fake data which is close to the real data can be synthesised. While early works focused on preventing low-quality generations, instability and model collapse problems, \emph{e.g.} WGANs \cite{35_WGAN, 36_ImprovedWGAN} and DCGANs \cite{15_DCGAN}, recent applications utilised various supervisions to control the generated data. For example, conditional GANs \cite{37_ContitionalGAN} can generate samples conforming to provided label information. Holding the assumption that the data is sampled from a low-dimensional manifold, GANs have been successfully applied to manuplating facial poses, ages, \emph{etc.,} by interpolating latent features on the manifold. Among these, GaitGAN \cite{GaitGAN17} and MGANs \cite{he2019multi} are proposed for cross-view gait recognition. Concretely, GaitGAN utilizes GAN to generate GEIs at side view and later improved in \cite{yu2019gaitganv2}, where a multi-loss strategy is employed to increase the inter-class distances. In MGANs, a new template named Period Energy Image (PEI) is defined and generated by their framework. However, compared with their methods, our method can 1) extract more discriminant view-invariant features, which is robust for large cross-view gait recognition tasks and 2) generate GEIs at controlled view angles rather than a single view (\emph{e.g.} side view), which is used as important evidences for identity recognition and authentication applications.


\section{Methodology}\label{sec3}

\begin{figure*}[h]
	\centering
	\includegraphics[width=.9\linewidth]{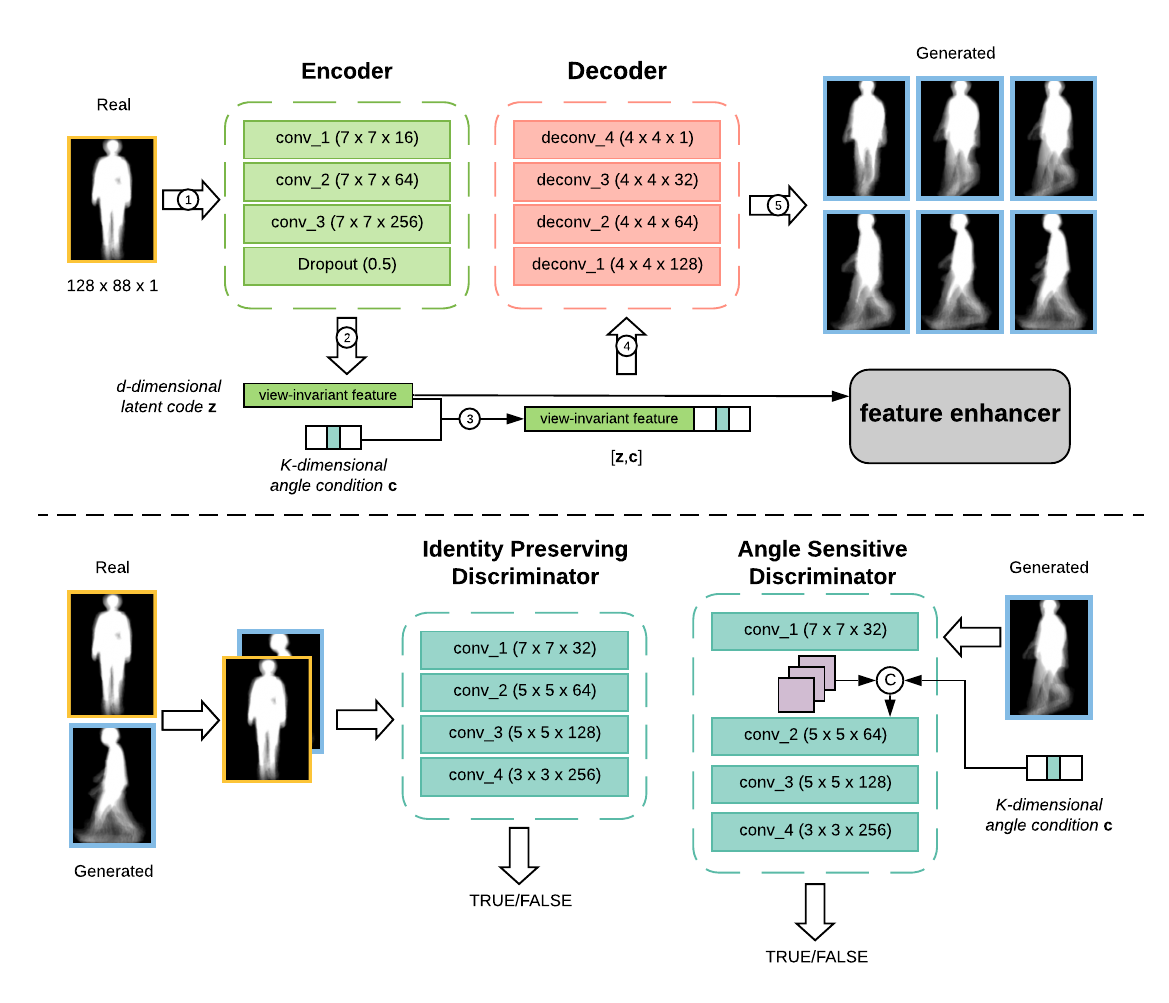}
	\caption{The proposed DiGGAN for cross-view gait recognition. The generator, which consists of an encoder and a decoder, is shown in the upper part of the figure. The view-invariant features of input GEIs are firstly extracted by the encoder (the progress is shown as void arrows 1 and 2 in the figure) and then concatenated (arrow 3) with a $K$-dimensional angle condition (one-hot vector). Subsequently, the concatenated vectors are fed into (void arrow 4) the decoder to generate (void arrow 5) the desired GEIs, which are expected to share the same identity with the input while at the conditioned view angles. Different with typical GANs, two discriminators (ID Discriminator and Angle Discriminator) are employed in DiGGAN. The ID Discriminator takes real and generated GEI pairs as input and checks whether they come from the same subject (if one of them is generated). The Angle Discriminator takes the generated image as well as the angle condition vector as input to validate if the generated image is at the expected view angle.}
	\label{fig:framework}
\end{figure*}

In this section, we describe the framework of the proposed DiGGAN and discuss the details of each component respectively. For a convenient discussion, in the rest of the paper, we use $\bm{x}_i^m$ to denote the GEI taken from the $i^{\text{th}}$ subject at angle $m$, thus $i \in \{1, 2,..., N\}$ and $m \in \{1, 2,..., K\}$, where $N$ is the number of subjects and $K$ is the number of views in the dataset.

Figure~\ref{fig:framework} illustrates an overview of the proposed framework. The proposed DiGGAN consists of two parts, a generator and dual discriminators. The generator is designed as a symmetric structure, in which the input GEIs are encoded as view-invariant features by the encoder, while the decoder recovers GEIs with different view angles from the concatenation of the view-invariant features and the angle condition. The dual discriminators are composed of an identity discriminator $D_{\text{id}}$ as well as an angle discriminator $D_{\text{angle}}$. 

The DiGGAN is proposed to transfer a GEI $\bm{x}_i^m$ at an arbitrary view angle $m$ to that at a controlled view angle $v$ while preserving the identity $i$, thus the desired output is $\bm{x}_i^v$ in the dataset. By generating GEIs at controlled view angles, the view-invariant features can be visualised so that the decision making process can be revealed. To achieve this, the input GEI $\bm{x}_i^m$ is firstly input into the encoder to extract view-invariant features $\bm{z}_i^m=E(\bm{x}_i^m)$. The view-invariant features are supposed to be close,  in the feature space, to those from the same identity, regardless of the view angles of the GEIs. Ideally, the view-invariant features contain no view angle information but the identity information only. Thereby to control the generated GEI's view angle, a $K$-dimensional one-hot encoded angle condition $\bm{c}\in\{0,1\}^K$ is then concatenated with the features and input into the decoder to generate the image $\hat{\bm{x}}_i^v = G(E(\bm{x}_i^m),\bm{c}_v)$, where $\bm{c}_v$ indicates $1$ is assigned to the $v^{\text{th}}$ entry of $\bm{c}$; and $0$ to elsewhere. To ensure the generated image preserving the identity information, the identity discriminator is employed here that takes a pair of images, \textit{e.g.} $(\bm{x}_i^m, \hat{\bm{x}}_i^v)$ as the input, and distinguishes whether these two images are from the same subject. Similarly, to constrain the generated image to be at the controlled angle, the angle discriminator is introduced to determine if the input image is matched with the input angle condition. By extracting view-invariant features to conduct cross-view gait recognitions and using them to transfer input GEIs to different view angles, the DiGGAN can be used to better understand the decision-making process.

During the testing phase, the nearest neighbour search is conducted among the probe's (\emph{i.e.} query data) view-invariant features and those of the gallery (\emph{i.e.} reference data) for identification and the verification is achieved by comparing the Euclidean distances between view-invariant features with a threshold. To further increase the discriminability of the view invariant features, a constraint as a local enhancer, for example, the triplet loss \cite{schroff2015facenet} that pushes mismatched GEI pairs away while pulling matched GEI pairs close, can be introduced. We also show that the proposed framework allows different local enhancers on the view-invariant feature space $\mathcal{Z}$, and comparisons on them will be given and discussed later in the experiments section. In the following of this section, the details of each component in the proposed framework will be discussed.

\subsection{Generator}
\subsubsection{Encoder}
The generators in typical GANs take a vector randomly sampled from a uniformed distribution as the input and generate various data mimicking the distribution they have learned from the realistic data. In DiGGAN, to jointly tackle the recognition as well as the evidence generating task, the generator is designed as a conditional auto-encoder where the conditional vector indicates the desired view angle of the generated GEI. The encoder here extracts features from GEIs for the recognition task and, at the same time, as the input of the decoder to generate view angle transferred GEIs as evidences.
The architecture of the proposed encoder is shown in the green dotted box in Fig.~\ref{fig:framework}. The input GEI is in the shape of $128\times88\times1$, where the last dimension indicates the number of channels of the image. The encoder consists of three convolution layers with kernel size of $7\times7$. The number of kernels increases from $16$ to $256$ with a multiplier of $4$. Each convolution layer is followed by a max pooling layer as well as a batch normalisation \cite{ioffe2015batch} layer except for the last layer, which is followed by a dropout \cite{JMLR:v15:srivastava14a} layer with ratio of $0.5$. Thereby the final feature maps are in the shape of $32\times22\times256$, and later on formed into the view-invariant features with a fixed dimension $d$.

\subsubsection{Decoder}
After obtaining the view-invariant features, a $K$-dimensional one-hot encoded angle condition $\bm{c}$ is concatenated with the features and input into the decoder to generate GEIs at the conditioned view angle. Beholding the superior performance of DCGAN \cite{radford2015unsupervised} on image and video generating, the decoder is designed in the fashion of deconvolutional networks. The architecture of proposed decoder can be found in red dotted box in Fig.~\ref{fig:framework}. The kernel size of each layer is empirically set as $4\times4$ and the number of kernels is set to half of that from the previous layer for the first three layers. For the last layer, the number of kernels is set to one to match the channel of GEI. It is worth noting that, as $88$, the width of the input GEI is not a multiplier of $16$, the stripe size of the last deconvolution layer is set as $(2, 1)$ instead of $(2, 2)$, which is used in the rest deconvolution layers. To keep the consistency of the activation function and maintain the nature of the GEI, ReLU is employed as the activation function for all the convolutional/deconvolutional layers thus constrains the pixel value within the range $[0,1]$.

\subsection{Identity Discriminator}
To ensure the view-invariant features to preserve the identity information, an intuitive idea is to validate if the generated GEI can be visibly recognised as the same person with the input. Therefore we employ an identity discriminator $D_{\text{id}}$ to compete with the generator. The $D_{\text{id}}$ is designed as a conditional discriminator that takes GEI pairs as input. If the GEI pairs are from the same identity and sampled from the real data (positive pair), the $D_{\text{id}}$ is expected to output the label as true, otherwise if one of GEI in the pair is generated, $D_{\text{id}}$ is supposed to output the label as false. Thus the objective function can be derived as:


\begin{equation}
\label{eq:1}
\begin{split}
\mathcal{L}_{\text{id}}=\underset{E,G}{\min}~\underset{D_{\text{id}}}{\max}~\mathbb{E}_{\bm{x}_i^m,\bm{x}_i^v\sim P_{\text{data}}} [\log D_{\text{id}}(\bm{x}_i^m,\bm{x}_i^v)&\\
+\log (1-D_{\text{id}}(\bm{x}_i^m,G(E(\bm{x}_i^m),\bm{c}_v)))]&\text{,}
\end{split}
\end{equation} where the $E$ denotes the encoder and $G$ denotes the decoder.
It is worth noting that, in practise we also construct negative pairs consisting of both real images but from different identities and expect the identity discriminator to output false. Learning from negative pairs rather than only from positive ones can strengthen the discriminator's ability in distinguishing identity information, thus the gradient can later on guide the encoder to capture more sufficient discriminative features for recognition. 

\subsection{Angle Discriminator}

One fact held in this work is that the GEIs are sampled from a low dimensional manifold \cite{he2019multi}, where the identity and angle change gradually and smoothly along respective dimensions. Thus we can easily manipulate the angle of the generated image by concatenating various angle vectors to the view-invariant features. Inheriting the similar idea of $D_{\text{id}}$, we employ an angle discriminator $D_{\text{angle}}$ to ensure the view angle of the generated GEI is matched with the angle condition. Mathematically, for the input GEI $\bm{x}_i^m$ and the angle condition $\bm{c}_v$, the angle discriminator can be trained by:
\begin{equation}
\begin{split}
\mathcal{L}_{\text{angle}} =\underset{E,G}{\min}~\underset{D_{\text{angle}}}{\max}~\mathbb{E}_{\bm{x}_i^m\sim P_{\text{data}}} [\log D_{\text{angle}}(\bm{x}_i^m, \bm{c}_m) &\\
 +\log(1-D_{\text{angle}}(G(E(\bm{x}_i^m),\bm{c}_v),\bm{c}_v))] &\text{.}
\end{split}
\end{equation}
In practise, the one-hot angle condition $\bm{c}$ is concatenated after the first convolutional layer 
in $D_{\text{angle}}$ to obtain a better performance inspiring by \cite{perarnau2016invertible}.

\subsection{Interchangeable Local Enhancer on View-invariant Features}
\begin{figure}[ht]
	\centering
	\includegraphics[width=.8\linewidth]{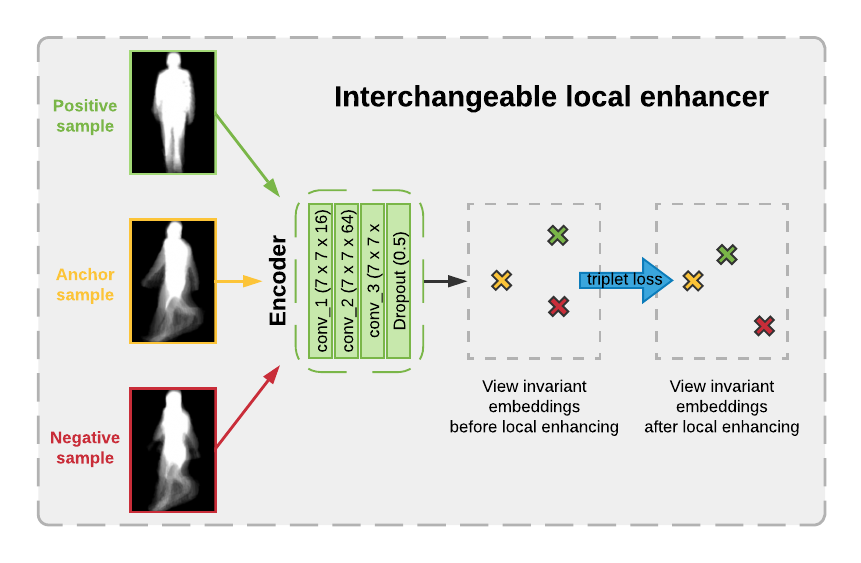}
	\caption{The constraints on view-invariant features: triplet loss as an example. The triplet loss can be employed to push the negative samples away from the anchor samples while pulling the positive samples closer.}
	\label{fig:triplet}
\end{figure}

Although the generated images can be directly used for gait recognition, \textit{e.g.} direct matching on the pixels \cite{YuShiqi}, searching on the latent feature space has been widely adapted by most of existing works \cite{PAMI17_CNN} \cite{he2019multi} for its higher performance and efficiency. To reduce the redundant information, we further refine the view-invariant features by proposing a local enhancer. It is worth noting that the local enhancing is a general concept and there may exist many implementations. Here we take the triplet loss \cite{schroff2015facenet} as an example to illustrate how the constraints enhance the discriminability of $\bm{z}$. As shown in Fig.~\ref{fig:triplet}, a triplet sample consists of an anchor, a positive and a negative sample, where the positive sample shares the same identity with the anchor while the negative one is different. To push the negative sample at least a margin $\delta$ further away from the anchor than the positive sample, for a triplet consisting of the anchor $\bm{x}_i^m$, the positive sample $\bm{x}_i^n$ and the negative sample $\bm{x}_j^l$, the objective function below is minimised to constrain their features $\bm{z}$:
\begin{equation}
\label{eq:3}
\begin{split}
	\mathcal{L}_{\text{le}} = \max(d(\bm{z}_i^m,\bm{z}_i^n)^2-d(\bm{z}_i^m,\bm{z}_j^l)^2+\delta,0) \text{,}
\end{split}
\end{equation}
where $d(\cdot,\cdot)$ can be ${L}_2$ distance. Note that here the view angle $l$ is not necessarily to be different with $m$ and $n$. As mentioned earlier, the local enhancer is not limited to triplet loss. Other constraints such as contrastive loss \textit{etc.} can also be used as the local enhancer, which will be compared in the experiment parts.

\subsection{Objective Function and Training Strategies}
\begin{figure}[]
	\includegraphics[scale=0.9, width=.8\linewidth]{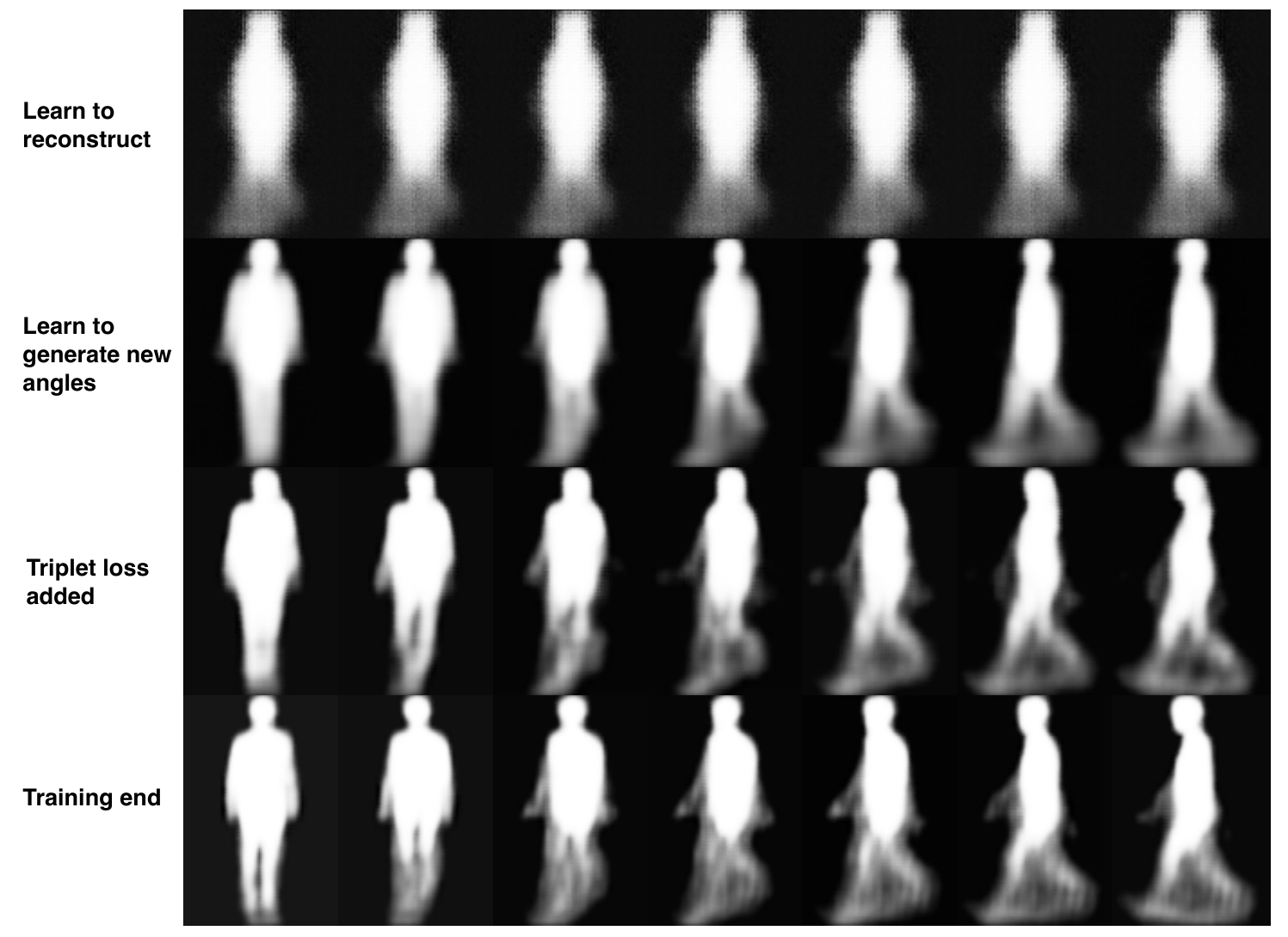}
	\caption{The generated images at different stages of the training process. First row: initial stage of the model. The model outputs averaged image. Second row: the model learns to generate images with new angles from $(\bm{x}_i^m,\bm{x}_i^v)$. Third row: after introducing local enhancer (triplet loss) into training, the model learns more identity details. Last row: model converges.}\label{fig:strategy}
\end{figure}

\noindent{\textbf{Reconstruction loss}} As one of the main motivations of this work is to generate evidences for identity recognition and authentication applications, where the GEIs' quality is a very important factor, besides the adversarial loss $\mathcal{L}_{\text{id}}$ and $\mathcal{L}_{\text{angle}}$ we introduce the pixel-wise reconstruction loss to further enhance the generated GEIs' quality. The reconstruction loss is defined as:
\begin{equation}
\label{eq:4}
	\mathcal{L}_{\text{rec}} = ||G(E(\bm{x}_i^m),\bm{c}_v),\bm{x}_i^v)||_1 \text{.}
\end{equation}
\noindent{\textbf{Overall objective function and training strategies}} Based on Eq.~\ref{eq:1} to \ref{eq:4}, we can define the overall objective function as follows:
\begin{equation}\label{eq:overall_loss}
\begin{split}
	\mathcal{L} = \lambda_1\mathcal{L}_{\text{le}} + \lambda_2\mathcal{L}_{\text{rec}} +\lambda_3\mathcal{L}_{\text{id}} +\lambda_4\mathcal{L}_{\text{angle}} \text{,}
\end{split}
\end{equation}
where $\lambda_{1-4}$ are the weights of each loss. Empirically, training such a model with multiple loss functions in Eq.~\ref{eq:overall_loss} is challenging thus always leads to poor results. To tackle this difficulty, we propose a step-by-step strategy for training. In the first step, we switch $\lambda_{1-3}$ off by setting them as zeros while leaving $\lambda_4$ on to train only the angle discriminator with artificial batches that are generated from the realistic GEIs. Specifically, we randomly sample $b$ GEIs from the training set to form a batch in which half of the images are assigned with wrong angle labels while the rest are assigned with the correct ones. Training with realistic images rather than generated ones helps the angle discriminator to be stronger to compete with the generator, avoiding the model collapse. We subsequently switch $\lambda_{2-3}$ on to train the network without the triplet loss in two sub stages. In the first sub stage, we set $\bm{x}_i^v=\bm{x}_i^m$ in Eq.~\ref{eq:1} and Eq.~\ref{eq:4}, which means the target GEI is set same as the input, therefore enables the network to learn to recover the input image first. Then in the second sub stage, we feed different images to teach the model to transfer the view angles. Finally, we take the local enhancer in and fine tune the whole network. Fig.~\ref{fig:strategy} shows the generated images at different stages of the training process. The model learns to generate averaged images at the initial stage. After that, with different images being fed into the network, the model learns to generate images of new angles. Finally the model learns to generate images with more details of the identity information with the local enhancer (triplet loss here).

\begin{table*}[ht]
	\centering
\caption{Rank-1 identification rate (\%) for all baselines in cooperative setting on OU-MVLP dataset.}
\label{tab:co_comparison}
\resizebox{1\textwidth}{!}{
\begin{tabular}{@{}llllllllllllllllllllll@{}}
\\\hline
\\
(a) VTM \cite{VTM06}   &  &       &      &  &      &  &      &  &      &  &  & \multicolumn{2}{l}{(b) GEINet \cite{GEINet16}} &       &      &  &      &  &      &  &      \\
           &  & Probe &      &  &      &  &      &  &      &  &  &                     &              & Probe &      &  &      &  &      &  &      \\ \cmidrule(lr){3-9} \cmidrule(lr){15-21}
Gallery    &  & 0     & 30   &  & 60   &  & 90   &  & Mean &  &  & Gallery             &              & 0     & 30   &  & 60   &  & 90   &  & Mean \\ \cmidrule(r){1-10} \cmidrule(l){13-22} 
0          &  & 68.8  & 0.5  &  & 0.2  &  & 0.1  &  & 17.4 &  &  & 0                   &              & 75.9  & 32.1 &  & 7.0  &  & 7.4  &  & 30.6 \\
30         &  & 0.7   & 82.2 &  & 2.1  &  & 0.8  &  & 21.4 &  &  & 30                  &              & 17.3  & 89.6 &  & 43.7 &  & 22.7 &  & 43.3 \\
60         &  & 0.3   & 3.2  &  & 77.6 &  & 5.4  &  & 21.6 &  &  & 60                  &              & 4.0   & 43.4 &  & 86.5 &  & 55.4 &  & 47.3 \\
90         &  & 0.2   & 1.1  &  & 4.2  &  & 80.9 &  & 21.6 &  &  & 90                  &              & 3.4   & 21.5 &  & 50.2 &  & 90.7 &  & 41.5 \\
Mean       &  & 17.5  & 21.7 &  & 21.0 &  & 21.8 &  & 20.5 &  &  & Mean                &              & 25.2  & 46.6 &  & 46.8 &  & 44.0 &  & 40.7 \\
           &  &       &      &  &      &  &      &  &      &  &  &                     &              &       &      &  &      &  &      &  &      \\
(c)Siamese \cite{Siamese16} &  &       &      &  &      &  &      &  &      &  &  & (d)CNN-MT \cite{PAMI17_CNN}               &              &       &      &  &      &  &      &  &      \\
           &  & Probe &      &  &      &  &      &  &      &  &  &                     &              & Probe &      &  &      &  &      &  &      \\ \cmidrule(lr){3-9} \cmidrule(lr){15-21}
Gallery    &  & 0     & 30   &  & 60   &  & 90   &  & Mean &  &  & Gallery             &              & 0     & 30   &  & 60   &  & 90   &  & Mean \\ \cmidrule(r){1-10} \cmidrule(l){13-22} 
0          &  & 52.7  & 23.7 &  & 11.1 &  & 11.3 &  & 24.7 &  &  & 0                   &              & 70.7  & 16.7 &  & 4.4  &  & 3.9  &  & 23.9 \\
30         &  & 18.4  & 78.6 &  & 32.6 &  & 27.6 &  & 39.3 &  &  & 30                  &              & 14.1  & 88.1 &  & 36.9 &  & 17.0 &  & 39.0 \\
60         &  & 8.0   & 33.5 &  & 76.1 &  & 39.6 &  & 39.3 &  &  & 60                  &              & 4.0   & 39.2 &  & 85.7 &  & 44.2 &  & 43.3 \\
90         &  & 7.9   & 26.5 &  & 36.5 &  & 82.1 &  & 38.2 &  &  & 90                  &              & 3.2   & 16.2 &  & 43.4 &  & 89.3 &  & 38.0 \\
Mean       &  & 21.8  & 40.6 &  & 39.1 &  & 40.1 &  & 35.4 &  &  & Mean                &              & 23.0  & 40.0 &  & 42.6 &  & 38.6 &  & 36.1 \\
           &  &       &      &  &      &  &      &  &      &  &  &                     &              &       &      &  &      &  &      &  &      \\
(e)CNN-LB \cite{PAMI17_CNN}     &  &       &      &  &      &  &      &  &      &  &  & (f)DM \cite{YuShiqi}               &              &       &      &  &      &  &      &  &      \\
           &  & Probe &      &  &      &  &      &  &      &  &  &                     &              & Probe &      &  &      &  &      &  &      \\ \cmidrule(lr){3-9} \cmidrule(lr){15-21}
Gallery    &  & 0     & 30   &  & 60   &  & 90   &  & Mean &  &  & Gallery             &              & 0     & 30   &  & 60   &  & 90   &  & Mean \\ \cmidrule(r){1-10} \cmidrule(l){13-22} 
0          &  & 74.4   & 16.5  &  & 3.5 &  & 2.8 &  & 24.3  &  &  & 0                   &              & 68.8   & 0.8 &  & 0.1 &  & 0.0 &  & 17.4 \\
30         &  & 13.6   & 89.3  &  & 36.0  &  & 16.2  &  & 38.8  &  &  & 30                  &              & 1.2  & 82.2  &  & 1.4 &  & 0.3 &  & 21.3 \\
60         &  & 2.9  & 36.2  &  & 88.4  &  & 44.7  &  & 43.0  &  &  & 60                  &              & 0.1  & 1.1 &  & 77.5  &  & 5.6 &  & 21.1 \\
90         &  & 2.2  & 14.0  &  & 41.2  &  & \textbf{91.7}  &  & 37.3  &  &  & 90                  &              & 0.0  & 0.2 &  & 4.1 &  & 80.9  &  & 21.3 \\
Mean       &  & 23.3   & 39.0  &  & 42.3  &  & 38.9  &  & 35.9  &  &  & Mean                &              & 17.5  & 21.1 &  & 20.8 &  & 21.7 &  & 20.3 \\
           &  &       &      &  &      &  &      &  &      &  &  &                     &              &       &      &  &      &  &      &  &      \\
(g)MGANs \cite{he2019multi}   &  &       &      &  &      &  &      &  &      &  &  & (h)DiGGAN(Ours)             &              &       &      &  &      &  &      &  &      \\
           &  & Probe &      &  &      &  &      &  &      &  &  &                     &              & Probe &      &  &      &  &      &  &      \\ \cmidrule(lr){3-9} \cmidrule(lr){15-21}
Gallery    &  & 0     & 30   &  & 60   &  & 90   &  & Mean &  &  & Gallery             &              & 0     & 30   &  & 60   &  & 90   &  & Mean \\ \cmidrule(r){1-10} \cmidrule(lr){13-22}  
0          &  & 72.0  & 9.6  &  & 6.8  &  & 2.4  &  & 22.7 &  &  & 0                   &              &   \textbf{79.0}    &  \textbf{62.1}   &  &  \textbf{46.5}  &  &  \textbf{47.7}   &  &  \textbf{58.8}   \\
30         &  & 9.4   & 83.2 &  & 30.3 &  & 10.7 &  & 33.4 &  &  & 30                  &              & \textbf{58.1}  & \textbf{89.8} &  & \textbf{64.8} &  & \textbf{58.5} &  & \textbf{67.8} \\
60         &  & 5.3   & 30.6 &  & 80.3 &  & 21.0 &  & 34.3 &  &  & 60                  &              & \textbf{44.1}  & \textbf{66.0} &  & \textbf{88.7} &  & \textbf{67.2} &  & \textbf{66.5} \\
90         &  & 2.1   & 12.0 &  & 22.0 &  & 85.9 &  & 30.5 &  &  & 90                  &              & \textbf{44.6}  & \textbf{58.9} &  & \textbf{66.0} &  & 90.0 &  & \textbf{64.8} \\
Mean       &  & 22.2  & 33.8 &  & 34.8 &  & 30.0 &  & 30.2 &  &  & Mean                &              &  \textbf{56.4}    &  \textbf{69.2}   &  &  \textbf{66.5}   &  &  \textbf{65.8}   &  &  \textbf{64.5} 
\\\hline  
\end{tabular}
}
\end{table*}

\section{Experiments}
In this section, we systematically evaluated our method on two most popular cross-view gait datasets, the OU-ISIR Gait Database, Multi-View Large Population Dataset (OU-MVLP) \cite{OU-MVLP2018} and CASIA-B \cite{casiab}. It is worth noting that the OU-LP \cite{OUISIR_LP} and USF \cite{USF_baseline} datasets are not used in this paper due to lack of large view changes. To evaluate the performance of our proposed method, we mainly focus on the following aspects:
\begin{itemize}
	\item Cross-view identification under the \textit{cooperative setting}, where the gallery has a uniform camera view angle.
	\item Cross-view identification under the \textit{uncooperative setting}, where the gallery contains unknown views.
	\item Effects of local enhancers on view invariant feature $\bm{z}$.
	\item Gait Evidence Generation
\end{itemize}

In the following, we will in turn introduce each of them. For all the experimental results, the DiGGAN model is trained, if not specified differently, with the triplet loss as the local enhancer on $\bm{z}$.

\subsection{Experimental Setup}
\noindent{\textbf{Datasets}} OU-MVLP is the world's largest cross-view gait dataset. It contains $10{,}307$ subjects ($5{,}114$ males and $5{,}193$ females with various ages, ranging from $2$ to $87$ years) and 14 different view angles \ang{0}, \ang{15}, \ang{30}, \ang{45}, \ang{60}, \ang{75}, \ang{90}, \ang{180}, \ang{195}, \ang{210}, \ang{225}, \ang{240}, \ang{255} and \ang{270}.
The subjects repeat forward and backward walking twice of each, such that two sequences are generated in each view.
The wearing conditions of subjects are various due to the collection process covering different seasons. The size-normalized GEIs used in this paper are $128 \times 88$ pixels.
Some examples from OU-MVLP dataset are illustrated in Fig. \ref{fig:dataset}.
CASIA-B is another widely used cross-view gait dataset that consists of $124$ subjects with $11$ different view angles range from \ang{0} to \ang{180} with an interval of \ang{18}. For each subject, there are six sequences of normal walking, two sequences with bags and two sequences with different clothes. 

\noindent{\textbf{Settings and hyper parameters}} For the experiments on OU-MVLP, we follow the settings in \cite{OU-MVLP2018}. The $10{,}307$ subjects in OU-MVLP dataset are split into two disjoint groups -- $5{,}154$ subjects for training our DiGGAN model and $5{,}153$ for testing, including probe and gallery. Similarly, for the CASIA-B dataset, we choose the first $62$ subjects for training and the rest $62$ subjects for testing with normal walking sequences. For the uncooperative settings, following \cite{OU-MVLP2018}, we randomly select one of the view angles for each test subject in gallery. The hyper parameters of proposed network are shown in Fig.~\ref{fig:framework} and the implementation code will be released if the work has been accepted. The dimension $d$ of the latent code $\bm{z}$ is set as $512$ for OU-MVLP and $128$ for CASIA-B; Empirically, the margins $\delta$ in the triplet loss and contrastive loss were set as $0.5$ and $1.0$ respectively for all the experiments.

\noindent{\textbf{Evaluation metrics}} Rank-1 identification rates (i.e., recognition accuracy) were used as the evaluation metric for the identification task and the equal error rates (EERs) were used to measure the performance on the verification task.

\subsection{Cross-view Gait Identification on Cooperative Setting}
\subsubsection{Experimental results on OU-MVLP}
Since two GEIs with \ang{180} view difference are mostly considered as those from the same-view pair based on perspective projection assumption \cite{makihara2006reference}, we focus on four typical view angels (\ang{0}, \ang{30}, \ang{60}, \ang{90}) in this section.
We compared our DiGGAN framework with some state-of-the-art baselines, including classical ones: direct matching (DM) \cite{YuShiqi}, VTM \cite{VTM06}, CNN-based methods: GEINet \cite{GEINet16}, Siamese \cite{Siamese16}, CNN-MT \cite{PAMI17_CNN}, CNN-LB \cite{PAMI17_CNN}, and the most recent GAN-based approach: MGANs \cite{he2019multi}. Note sequence-based (silhouette) methods are not compared, which were considered to be computationally expensive and sensitive to segmentation errors.
In the cooperative mode, the rank-1 identification rates of all four view angles are reported in Table \ref{tab:co_comparison}, from where we can see:
\begin{itemize}
  \item Our method outperforms other methods substantially on cross-view gait identification tasks. Our overall rank-1 accuracy is $64.5\%$, and that is $23.8\%$ higher than the second best GEINet.
  \item Our method is more robust on cross-view gait identification. In this cooperative mode, accuracies decrease w.r.t. increasing view angles differences, and they are less significant when compared with other algorithms. Our DiGGAN can yield very competitive performances even when the view difference is $\ang{90}$, which indicates our method can extract robust view-invariant features.
  \item Most of the methods suffered from gallery in view \ang{0}, yet our DiGGAN can still achieve a reasonable accuracy of $58.8\%$, much higher than the second best.
\end{itemize}
In Table \ref{tab:cross_view}, we also report the average rank-1 accuracies on cross-view gait identification excluding the identical views (between probe and gallery).
We can see other algorithms do not generalise well in this large-scale cross-view gait recognition evaluation, while our DiGGAN can still remain competitive results.

\begin{table}[h]
	\caption{Average rank-1 identification rates (\%) under Probe \ang{0}, \ang{30}, \ang{60} and \ang{90} excluding identical view (cooperative mode) on OU-MVLP dataset.}
	\label{tab:cross_view}
	\resizebox{0.5\textwidth}{!}{
		\begin{tabular}{llllllllll}
			\hline
			&  & Probe &      &  &      &  &      &  &      \\ \cline{3-10}
			Method  &  & \ang{0}     & \ang{30}   &  & \ang{60}   &  & \ang{90}   &  & Mean \\ \hline
			VTM \cite{VTM06}     &  & 0.4   & 1.6  &  & 2.2  &  & 2.1  &  & 1.6  \\
			GEINet \cite{GEINet16}  &  & 8.2   & 32.3 &  & 33.6 &  & 33.6 &  & 26.9 \\
			Siamese \cite{Siamese16} &  & 11.4  & 27.9 &  & 26.7 &  & 26.2 &  & 23.1 \\
			CNN-MT \cite{PAMI17_CNN}  &  & 7.1   & 24.0 &  & 28.2 &  & 21.7 &  & 20.3 \\
			CNN-LB \cite{PAMI17_CNN}  &  & 6.2   & 22.2 &  & 26.9 &  & 21.2 &  & 19.1 \\
			DM \cite{YuShiqi}      &  & 0.4   & 0.7  &  & 1.9  &  & 2.0  &  & 1.3  \\
			MGANs \cite{he2019multi}   &  & 5.6   & 17.4 &  & 19.7 &  & 11.4 &  & 13.5 \\
			DiGGAN(ours)    &  & \textbf{48.9}  & \textbf{62.3} &  & \textbf{59.1} &  & \textbf{57.8} &  & \textbf{57.0} \\ \hline
		\end{tabular}
	}
\end{table}

\subsubsection{Experimental results on CASIA-B} CASIA-B is a relative small dataset. We evaluated our model and the average recognition accuracies on CASIA-B are reported in Table~\ref{tab:casia-b}. The comparison is conducted under the probe views at \ang{54}, \ang{90} and \ang{126} and with several methods such as VTM \cite{VTM_SVR}, C3A \cite{CCA}, ViDP \cite{ViDP}, CNN-MT \cite{PAMI17_CNN} and MGANs \cite{he2019multi}. The results show that our method yields the competitive performance under probe \ang{54} while getting significant improvements under probe \ang{90} and \ang{124}, which indicates our framework works well on small scale datasets.

\begin{table}[h]
	\caption{Average rank-1 identification rates (\%) under Probe \ang{54}, \ang{90} and \ang{126} excluding identical view (cooperative mode) on CASIA-B dataset.}
	\label{tab:casia-b}
	\centering
	\resizebox{0.5\textwidth}{!}{
	\begin{tabular}{llllllll}
		\hline
		       					&  & Probe         &               &  &               &  &        \\ \cline{3-8} 
		Method 					&  & \ang{54}      & \ang{90}      &  & \ang{126}     &  & Mean   \\ \hline
		VTM \cite{VTM06}   		&  & 55.0          & 46.0          &  & 54.0          &  & 51.0   \\
		C3A \cite{CCA}    		&  & 75.7          & 63.7          &  & 74.8          &  & 71.4   \\
		ViDP \cite{ViDP}   		&  & 64.2          & 60.4          &  & 65.0          &  & 63.2   \\
		CNN-MT \cite{PAMI17_CNN}   	&  & \textbf{94.6}          & 88.3          &  & 93.8          &  & 92.2   \\
		MGANs \cite{he2019multi}  &  & 84.2          & 72.3          &  & 83.0          &  & 79.8   \\
		DiGGAN(ours)   &  &  94.4 & \textbf{91.2} &  & \textbf{93.9} &  & \textbf{93.2} \\ \hline
	\end{tabular}
	}
\end{table}

\subsubsection{Cross dataset evaluation}
In this section, we evaluated the generalisation ability of our model. We trained three models, among which the first model ($M_{\text{O}}$) is trained on OU-MVLP dataset only, the second model ($M_{\text{C}}$) is trained on CASIA-B dataset and the last model ($M_{\text{O+C}}$) is first trained on OU-MVLP and then fine-tuned on CASIA-B. We report the average rank $1$ identification rates of each model on the $62$ subjects in CASIA-B's test set, and the results are shown in Table \ref{tab:cross-dataset}. We can see that the model $M_{\text{O}}$ trained on OU-MVLP yields a promising identification rate on CASIA-B dataset. We can also find that pre-training on OU-MVLP dataset helps the model $M_{\text{O+C}}$ to achieve a slightly improvement on the performance compared to the $M_\text{C}$ because of its massive number of training samples. However, we noticed that $M_{\text{O+C}}$ does not benefit much from a large pretrain set. A possible reason is that the view angles as well as the nationalities (Chinese, Japanese) of the subjects in OU-MVLP and CASIA-B are different. Nevertheless, the experimental results suggest it is not harmful to use the large OU-MVLP for representation learning. In fact, based on the learned representation, even without local fine tuning, our model $M_{\text{O}}$ can outperform all the existing methods except the CNN-MT \cite{PAMI17_CNN}, which shows our framework has a very strong generalisation ability.

\begin{table}[h]
	\caption{Cross dataset evaluation on CASIA-B dataset: Average rank-1 identification rates (\%) under Probe \ang{54},\ang{90} and \ang{126} excluding identical view (cooperative mode).}
	\label{tab:cross-dataset}
	\centering
		\begin{tabular}{llllllllll}
		\hline
		&  & Probe &      &  &      &  &      &  &      \\ \cline{3-10}
		Model  &  & \ang{54}     & \ang{90}   &  & \ang{126}      &  & Mean \\ \hline
		$M_{\text{O}}$    &  & 86.2   & 82.2  &  & 84.7  &  & 84.4    \\
		$M_{\text{C}}$  &  & 94.4   & 91.2 &  & 93.9 &  & 93.2  \\
		$M_{\text{O+C}}$ &  & \textbf{94.6}  & \textbf{91.3} &  & \textbf{93.9} &  & \textbf{93.3}  \\ \hline
	\end{tabular}
\end{table}

\subsection{Cross-view Gait Identification on Uncooperative Setting}
\subsubsection{Uncooperative setting results} Compared with cooperative mode, this scenario is more challenging since the gallery contains non-uniform views.
Following the settings in \cite{OU-MVLP2018}, we randomly select one from the $14$ view angles for each test subject in gallery.
The experimental results are shown in Table~\ref{tab:unco_comparison}, our model significantly outperforms state-of-the-art approaches.

\begin{table}[h]
	\caption{Rank-1 identification rate (\%) for all baselines in uncooperative setting on OU-MVLP dataset.}
	\label{tab:unco_comparison}
		\begin{tabular}{llllllllll}
			\hline
			&  & Probe         &               &  &               &  &               &  &               \\ \cline{3-8}
			Method     &  &\ang{0}     & \ang{30}   &  & \ang{60}   &  & \ang{90}          &  & Mean          \\ \hline
			GEINet \cite{GEINet16} &  & 15.7          & 41.0          &  & 39.7          &  & 39.5          &  & 34.0          \\
			Siamese \cite{Siamese16}    &  & 15.6          & 36.2          &  & 33.1          &  & 36.5          &  & 30.3          \\
			CNN-LB \cite{PAMI17_CNN}         &  & 14.2          & 32.7          &  & 32.3          &  & 34.6          &  & 28.5          \\
			CNN-MT \cite{PAMI17_CNN}         &  & 11.1          & 31.5          &  & 31.1          &  & 29.8          &  & 25.9          \\
			DM \cite{YuShiqi}         &  & 7.1           & 7.4           &  & 7.5           &  & 9.7           &  & 7.9           \\ \hline
			DiGGAN(ours)       &  & \textbf{30.8} & \textbf{43.6} &  & \textbf{41.3} &  & \textbf{42.5} &  & \textbf{39.6} \\ \hline
		\end{tabular}
\end{table}

\subsubsection{Performance on small-scale gallery} 
\begin{figure}[t]
\begin{center}
   \includegraphics[width=0.99\linewidth]{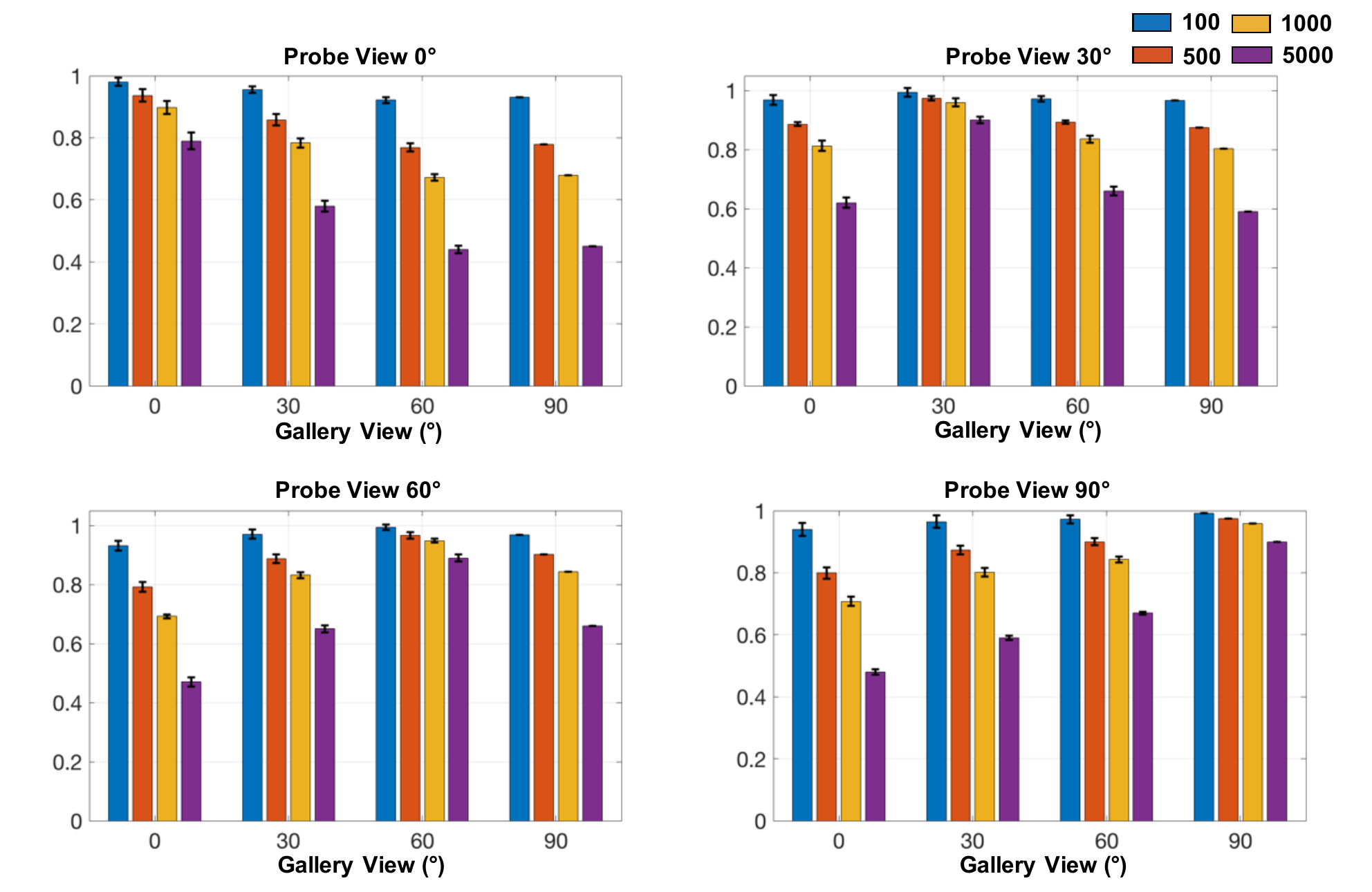}
\end{center}
\vspace{-2ex}
   \caption{Performance \textit{w.r.t.} the size of gallery (in cooperative mode) on OU-MVLP.}
   \vspace{-2ex}
\label{fig:subset}
\end{figure}

In many realistic applications, such as indoor office with limited number of subjects, the gallery size can be smaller.
In Fig. \ref{fig:subset}, we can see the performance tends to be higher with smaller gallery.
At $100$-identity scale, the accuracies under all views exceed 90\%, which is in line with the experimental results on the small-scale CASIA-B. Given the high performance, our model has many potential industrial values.

%

\subsection{Effects of Local Enhancers on View-invariant Features}
To better understand our proposed model, this section provides detailed discussions and verifies some key statements in our methodology. 



\subsubsection{Effects on cross-view gait identification} We first study the effects of the local enhancers on cross-view gait identification task. We respectively set the local enhancer in our framework as the triplet loss (mentioned in Eq.~\ref{eq:3}) and the contrastive loss, which is written as:
\begin{equation}
	\mathcal{L}_{\text{ctr}} = yd(\bm{z}_\alpha,\bm{z}_\beta)^2+(1-y)\max(\delta-d(\bm{z}_\alpha,\bm{z}_\beta),0)^2 \text{,}
\end{equation}
where $\bm{z}_\alpha$ and $\bm{z}_\beta$ are view-invariant features extracted from any two GEIs, (note here $\bm{z}_\alpha$ does not necessarily correspond to the $\alpha^{\text{th}}$ identity) and $y$ equals $1$ if $\bm{z}_\alpha$ and $\bm{z}_\beta$ are from the same identity and $0$ otherwise. Similar to the triplet loss, $d(\cdot,\cdot)$ here denotes ${L}_2$ distance. We evaluated the performances in cooperative setting at four typical views in gallery and reported the average rank-1 identification rate in Table.~\ref{tab:triplet}. We can see that both the contrastive loss and the triplet loss can bring us some improvements on the performances compared with the other results in Table.~\ref{tab:co_comparison}. Additionally, the DiGGAN with contrastive loss outperforms the Siamese \cite{Siamese16}, which has also employed contrastive loss to extract gait features, and we conclude that the improvement is brought by the dual discriminators in our framework. Comparing contrastive loss with the triplet loss, we can see that the triplet loss outperforms the contrastive loss significantly. Because contrastive loss is based on the assumption that the distribution of the features within each class is the same, which is not always true, especially in human's gaits. In the future, other metric learning methods can be explored to further boost the performance.

\begin{table}[h]
\caption{Effects on local enhancers: average rank-1 identification rates (\%) on OU-MVLP dataset. Gallery views are at \ang{0}, \ang{30}, \ang{60} and \ang{90}.}
\label{tab:triplet}
	\begin{tabular}{llllllllll}
		\cline{1-8}
		&   Probe                    &                           &  &                           &  &                           &  &      \\ \cline{2-8} 
		Method              			    & \ang{0}                  & \ang{30}                  &  & \ang{60}                 &  & \ang{90}                     & Mean \\ \cline{1-8}
		DiGGAN(contrastive) 			     & 27.7                     & 47.0                      &  & 44.2                      &  & 45.8                        & 41.2 \\
		DiGGAN(triplet)       & \textbf{56.4}                     & \textbf{69.2}                      &  & \textbf{66.5}                      &  & \textbf{65.8}                        & \textbf{64.5} \\ \cline{1-8}
	\end{tabular}
\end{table}

\begin{figure*}[ht]
	\centering
	\includegraphics[width=0.95\textwidth]{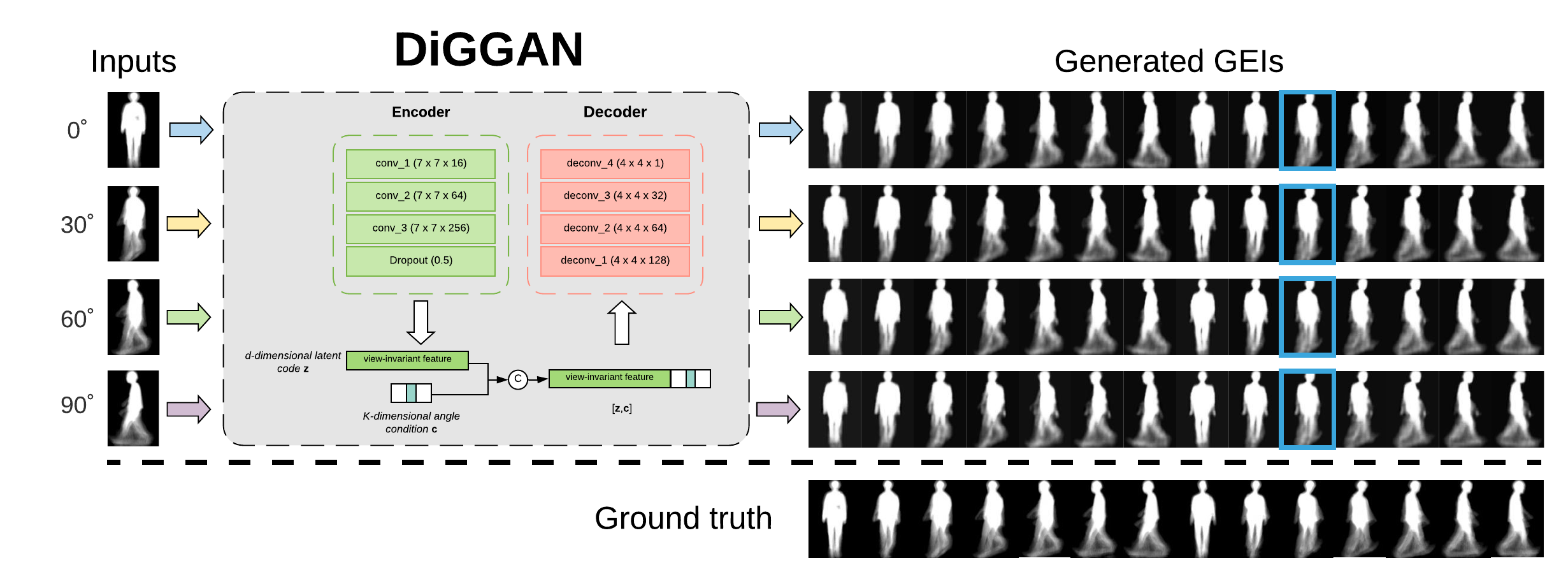}
	\caption{Illustration of the capability of any-to-any view evidence generation of DiGGAN. The inputs shown in the figure are at \ang{0}, \ang{30}, \ang{60} and \ang{90}. The corresponding generated GEIs are shown in the right side. The ground truth GEIs are shown in the bottom.}\label{evidence}
\end{figure*}



\begin{figure}
	\begin{center}
		\includegraphics[width=0.6\textwidth]{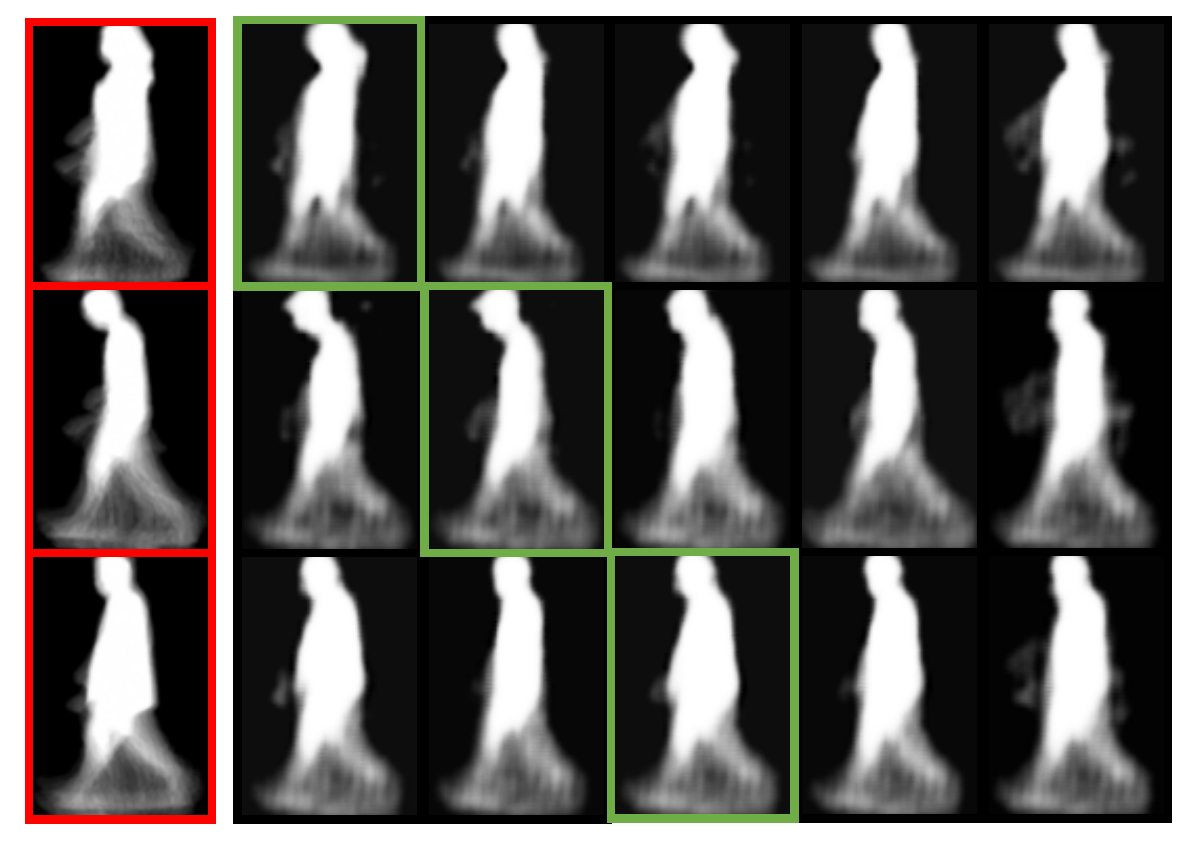}
		\caption{Qualitative analysis of the evidence generation. The first column marked with red box illustrates three different GEIs in the probe. The rest five images in each row are generated GEIs based on the five most similar features.\label{evidence_3}}
	\end{center}
\end{figure}



\subsubsection{Effects on cross-view gait verification}
We also evaluated the proposed model with triplet loss and contrastive loss on cross-view gait verification task and compared our method with the state-of-the-art. The experimental results on gait verification at four typical angular differences (between probe and gallery) are given in Table.~\ref{tab:EER}. We can see the DiGGAN with triplet loss yields the best EERs on angular difference at \ang{60} and \ang{90}. However, the best performance on angular difference at \ang{0} and \ang{30} comes from CNN-MT \cite{PAMI17_CNN} and 2in+diff \cite{takemura2017input} respectively. This is mainly because they employ either a feature level \cite{PAMI17_CNN} or model level \cite{takemura2017input} fusion, and these fusion strategies will be explored in future work.

\begin{table}[h]
	\caption{Effects on local enhancers: EERs (\%) of cross-view gait verification on OU-MVLP dataset. Angular difference between probe and gallery are at \ang{0}, \ang{30}, \ang{60} and \ang{90}.}
	\label{tab:EER}
	\begin{center}
		\begin{tabular}{llllll}
			\hline
			& \multicolumn{5}{l}{Angular Difference} \\ \cline{2-6} 
			Method              & \ang{0}     & \ang{30}   &   \ang{60}   &  \ang{90}    & Mean  \\ \hline
			DM \cite{yu2006framework}                 & 6.5   & 25.2   & 41.4  & 46.2  & 27.2  \\
			VTM \cite{VTM06}               & 6.5   & 26.8   & 34.2  & 38.5  & 25.0  \\
			GEINet \cite{GEINet16}              & 2.4   & 5.9    & 12.7  & 17.2  & 8.1   \\
			CNN-LB \cite{PAMI17_CNN}              & 1.0   & 3.3    & 6.7   & 9.3   & 4.3   \\
			CNN-MT \cite{PAMI17_CNN}              & \textbf{0.9}   & 2.5    & 5.2   & 7.0   & 3.3   \\
			2in+diff \cite{takemura2017input}            & 1.0   & \textbf{2.0}    & 3.4   & 4.2   & \textbf{2.4}   \\ \hline
			DiGGAN(contrastive) & 1.3   & 2.3    & 3.5   & 4.4   & 2.6   \\
			DiGGAN(triplet)     & 1.2   & 2.2    & \textbf{3.3}   & \textbf{4.1}   & \textbf{2.4}   \\ \hline
		\end{tabular}
	\end{center}
\end{table}

\subsection{Gait Evidence Generation}
\subsubsection{Any-to-any view gait evidence generation}
One of the advantages of our proposed model is that we can generate GEIs from a certain view angle to all target angles (existing in training sets). Such an extension helps the decision makers to understand when the identification and verification are based on the view-invariant features.

To explore the capability of our model on gait evidence generation, we conducted an any-to-any GEI generation experiment, where we took GEIs at four typical views (\ang{0}, \ang{30}, \ang{60} and \ang{90}) as the inputs to generte GEIs at all $14$ views (\ang{0} - \ang{90}, \ang{180} - \ang{270}). The generated GEIs as well as the corresponding ground truth are shown in Fig.~\ref{evidence}. From each pathway we can see that the generated GEIs have successfully preserved the important details of the ground truth GEIs, including the limbs posture, neck posture, \emph{etc.} Additionally, looking at the generated GEIs at the same view angle (\emph{e.g.} bounded with blue boxes) across the four pathways, we can also find that the variances among them are very small, which suggests the view information has been disentangled in the proposed view-invariant features.

\subsubsection{Feature-evidence consistencies}
The effectiveness of our generated evidences can be reflected by the consistencies between feature level similarity and evidence level similarity. If the generated evidences (GEIs) recovered from the most similar view-invariant features are consistently similar, it is confident to say the generated evidences are convincing. To reveal this, we picked the GEIs of three individuals at view \ang{90} from the probe (shown with red boxes in Fig.~\ref{evidence_3}) and extracted the view-invariant features. Then, we performed a nearest neighbour search and selected the top-5 ranked features from the gallery to generate GEIs. In Fig.~\ref{evidence_3}, the green boxed GEIs are generated with the features extracted from correct individuals. We can see there is a good consistency between the feature level and evidence level similarity of our model. Especially in the first row, we can find the generated evidence shares a great similarity with the probe. These gait evidences are able to offer us an opportunity to double check: if one finds the gait evidences generated from the top ranked features are not similar with the probe, re-decision with the helps from human experts can be introduced.

\subsubsection{Discussions for evidence generation}
The any-to-any view gait evidence generation experiment has shown the ability of our proposed model in generating GEIs at different views, and the feature-evidence consistencies experiment has revealed the validity of the generated GEIs --- the visual similarities between generated GEIs and original real GEIs are consistent with the similarities between their features. These two experiments together have proved the proposed DiGGAN is able to generate convincing GEIs as important evidences. For example, when the criminal suspect's GEI left at crime scence is different with his GEI stored at police database, the DiGGAN can be used to verify if the suspect's GEI is the same with the GEI in the database, and at the same time convert the GEI to the same view angle to that in database to let the officers to check with their domain knowledges. However there are still limitations of our work. The current version of DiGGAN is limited to generate GEIs only at view angles that have shown in training dataset. If the required view angles have never shown up, then it is not feasible for DiGGAN to generate them. We leave this feature as our future work.  


\section{Conclusion}

This paper proposed a Discriminant Gait Generative Adversarial Network (DiGGAN) that tackles cross-view gait recognition problems. Systematic experiments have been conducted on two most popular cross view gait recognition datasets (\emph{i.e.} OU-MVLP and CASIA-B), and the experimental results show the proposed DiGGAN outperforms the state-of-the-art methods on both gait identification and verification tasks. Cross dataset evaluation on CASIA-B has also been conducted and shows that our model has reasonable generalisation ability. Moreover, a series of qualitative experiments illustrate the DiGGAN is able to transfer the input GEI to different view angles, which suggests these generated GEIs can be used as evidences in security and privacy applications. Additionally, we discussed the limitation of current DiGGAN in generating GEIs at unseen view angles, which is regarded as the future direction of our research.


%
\bibliographystyle{ACM-Reference-Format}
\bibliography{reference}

\appendix

\end{document}